# Superpixel Perception Graph Neural Network for Intelligent Defect Detection of Aero-engine Blade

Hongbing Shang, Qixiu Yang, Chuang Sun*, Xuefeng Chen, Ruqiang Yan

School of Mechanical Engineering, Xi'an Jiaotong University, Xi'an, Shaanxi 710049, China

## Abstract

Aero-engine is the core component of aircraft and other spacecraft. The high-speed rotating blades provide power by sucking in air and fully combusting, and various defects will inevitably occur, threatening the operation safety of aero-engine. Therefore, regular inspections are essential for such a complex system. However, existing traditional technology which is borescope inspection is labor-intensive, time-consuming, and experience-dependent. To endow this technology with intelligence, a novel superpixel perception graph neural network (SPGNN) is proposed by utilizing a multi-stage graph convolutional network (MSGCN) for feature extraction and superpixel perception region proposal network (SPRPN) for region proposal. First, to capture complex and irregular textures, the images are transformed into a series of patches, to obtain their graph representations. Then, MSGCN composed of several GCN blocks extracts graph structure features and performs graph information processing at graph level. Last but not least, the SPRPN is proposed to generate perceptual bounding boxes by fusing graph representation features and superpixel perception features. Therefore, the proposed SPGNN always implements feature extraction and information transmission at the graph level in the whole SPGNN pipeline, to alleviate the reduction of receptive field and information loss. To verify the effectiveness of SPGNN, we construct a simulated blade dataset with 3000 images. A public aluminum dataset is also used to validate the performances of different methods. The experimental results demonstrate that the proposed SPGNN has superior performance compared with the state-of-the-art methods.

---

* Corresponding author. E-mail address: ch.sun@xjtu.edu.cn (Chuang Sun)

# 1 Introduction

With the advent of industry 4.0 and the era of big data, Prognostic and Health Management (PHM) has been becoming increasingly critical for industrial complex systems (ICSs), such as aeroengine [1] and other industrial products [2-5]. Efficient and reliable health management systems are an integral component of ICS due to their capabilities of monitoring, diagnosis, and prognostic [6-8]. In general, PHM significantly improves equipment reliability, reduces downtime maintenance costs, and improves economic efficiency. Therefore, related research in this field has been in full swing in recent years, and various monitoring and diagnostic methods emerge in an endless stream.

Aero-engine is the core component of aircraft and other spacecraft. As a typical ICS operating in harsh working conditions, regular inspection and maintenance are critical to its operation safety [9, 10]. In aero-engine, the high-speed rotating blades provide power by sucking in air and fully combusting, and various defects will inevitably occur, threatening the operation safety of aero-engine. Therefore, efficient and reliable in-situ blade damage detection is a vital link to aero-engine planned ground maintenance. In-situ blade damage detection refers to level-by-level and one-by-one blade visual recognition by the naked eye with the help of borescope on the tarmac or hangar without disassembling the aero-engine.

As a mainstream technology, borescope inspection plays a significant role in in-situ inspection and maintenance of aero-engine [11, 12]. Specifically, several experienced inspectors circled the aero-engine with borescope and maintenance manual before takeoff or after landing. The inspection process is cumbersome: 1) Open the reserved peephole on the aero-engine. 2) Insert the probe of the borescope into the aero-engine. 3) Shake the rotor blades of the entire engine through the handle. 4) Stare at the screen of the hole detector and scan the blades one by one with the naked eye. 5) Identify the defect against the maintenance manual. In addition, it takes a long time to train a qualified inspector, and the technical level of different inspectors is uneven, resulting in subjective evaluation results and formidable difficulty.

For the issue of borescope image processing, scholars have carried out much research by using traditional image processing technologies. For example, Aust et al. [13] developed an automatic defect detection system by designing a heuristic algorithm, and grey-scaling, filtering, and feature extraction are also adopted to detect and locate defects. Guo et al. [14] proposed a weighted morphology image segmentation method by combining weighted structural element and edge operator. Holak and Obrocki [15] presented an image processing technology by introducing standard deviation image filtering and fractal dimension processing. Besides, Zuo [16] reviewed the traditional image processing technology for aero-engine damage detection including edge detection, threshold segmentation, and wavelet transform. However, these methods have suboptimal detection performance due to hand-crafted features, elusive prior, and unacceptable efficiency. In addition, the inability to train and test end-to-end limits the practical application of these methods.

In recent years, deep learning (DL) based defect detection methods have become dominant technologies in academia and industry. Li et al. [17] and Fink et al. [18] reviewed deep transfer learning and deep learning methods in industrial application, and found that CNNs have become the de-facto architecture for various visual inspection methods. Various dazzling DL-based algorithms are proposed to achieve intelligent borescope inspection. For example, Wong et al. [19] presented a real-time jet engine blade damage assessment method by training an improved instance segmentation network, their method directly extracts defect feature and lacks defect characteristics analysis. Ren et al. [20] proposed a lightweight aircraft defect detection model which is an ensemble of multi convolutional neural networks (CNNs), however, the model only classifies defect types without being able to detect and quantify defects. Kähler et al. [21] designed a CNN-based autoencoder to reconstruct defect images for anomaly detection of aircraft components, but they treat visual inspection as an anomaly detection task, rather than recognizing or detecting defects. Svensén et al. [22] adopted CNN to process borescope images from GE Aviation and showed excellent experimental results in the classification of engine parts, but their method lacks further borescope image analysis about location and extent of defect. Bian et al. [23] proposed a hierarchical fully CNN architecture for the inspection of aircraft engine blades, their method is not tailor-made for aircraft engine inspection instead of utilizing multi-scale convolutional operator in a plain way.

In other industrial applications, Qu et al. [24] explored a multiscale feature fusion strategy for pavement crack detection, their method still replies on CNN-based feature extraction network. Hu et al. [25] added a sequence-PCA layer and attention module to provide enhanced semantic information and obtain better binary classification performance. Zou et al. [26]

proposed an end-to-end CNN to extract hierarchical features of crack for crack segmentation. However, their methods are not suit for multi-class blade defect detection in borescope image processing. Chen et al. [27] reported a cascade CNN-based defect detection algorithm in a coarse-to-fine manner, but the three stages of their method are separated. Generative adversarial networks (GANs) are also used for data augmentation [28] and semantic segmentation [29] in defect detection. In addition, CNN-based autoencoders are developed for industrial defect detection [30]. Overall, these methods almost adopt CNNs and do not consider the characteristics of images to develop environment adaptation algorithms.

Due to the properties of locality, sparse connections, and parameter sharing, CNNs have swept various computer vision tasks. In the final analysis, the reason for this phenomenon comes down to the powerful ability of CNN in image feature extraction. Although CNNs have been shown to have great advantages in defect detection, this network architecture may be not suitable for the scene of borescope inspection due to specific engineering application and realistic borescope images. We summarize the reasons for this phenomenon as follows: 1) *Intrinsic Limitation of CNNs*: The regular locality and limited receptive field of CNNs make CNN-based methods extract features by continuously stacking convolutional layers, resulting in utilizing long-range contextual information inefficiently. Another fatal flaw is that the convolution operator processes image patches in regular grids, which is not suitable for irregular defect feature extraction. Therefore, seeking a suitable and efficient feature extraction method is crucial for defect detection. 2) *Irregular and Complex Textures*: There are various defect types on the borescope images, and the size of these defects varies dramatically. Even though there are differences between defects of the same type. We elaborate on the various types of blade defects in Table I. Therefore, capturing irregular and complex textures requires efficient and accurate defect feature extraction. 3) *Harsh Working Condition*: The closed environment makes the borescope image quality unsatisfactory, and the illumination influence and dust pollution also make intelligent methods difficult to extract defect features. The interweaving fusion of defects, blades, and background increases the difficulty of defect detection. Therefore, the defects can be effectively focused on if the perception of the defect region is achieved.

Table I. Typical damage modes and characteristics of aero-engine blades

| Mode | Crack | Burned | Broken | Nick | Overheated | Dent | Tears | Tip curl | Burned through |
|---|---|---|---|---|---|---|---|---|---|
| Images | 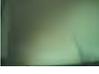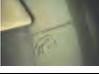 | 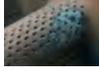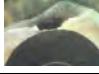 | 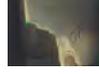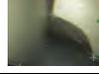 | 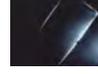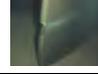 | 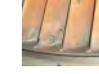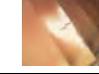 | 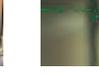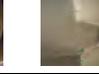 | 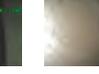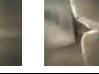 | 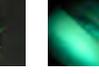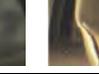 | 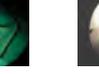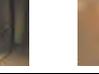 |
| art | C1/T/C2 | T/C2 | C1/T | C1/T | T/C2 | C1 | C1 | C1 | C1/T |

where C1 denotes compressor, C2 denotes combustor, and T denotes turbine.

To address the issues mentioned above, we propose a novel superpixel perception graph neural network (SPGNN) by utilizing a multi-stage graph convolutional network (MSGCN) for feature extraction and superpixel perception region proposal network (SPRPN) for region proposal. In SPGNN, borescope images are processed into a series of graph patches. Thus, the graph data is constructed by viewing each patch features and the relations between each node and its neighbors as nodes and edges, respectively. The advantages of this processing method are that adaptive feature extraction and information aggregation are feasible when GNNs are utilized to process graph data, overcoming the limitation of the local receptive field of CNNs. Then, MSGCN extracts multi-scale features at graph level instead of regular grid, so it generates hierarchical feature maps with rich representations for defect detection of irregular and complex textures. Besides, by processing features from superpixel segmentation through a learnable GCN, SPRPN generates perceptual bounding boxes that focus on defect regions and eliminate background distractions. We summarize the contributions of our work as follows:

1) This work formulates local grid processing and borescope image feature extraction as graph structure construction and graph information aggregation, respectively. Thus, MSGCN is naturally applied on a series of graph patches for defect feature extraction from a fresh graph representation learning perspective.

2) The proposed SPRPN as a graph RPN integrates graph representation features from MSGCN and superpixel perception features which are extracted by applying a learnable GCN and the mask strategy on superpixel graphs in SPRPN. Therefore, the feature extraction and information transmission are implemented at the graph level in the "feature extraction-region proposal" graph representation learning pipeline, to alleviate the reduction of receptive field and information loss.

3) Extensive experiments are carried out on public and homemade datasets to validate the effectiveness of the proposed SPGNN. The results show that SPGNN has superior performance. Ablation experiments and visualization results are presented to demonstrate the effects of MSGCN and SPRPN. Parameter sensitivity is discussed to understand SPGNN comprehensively and thoroughly.

## 2 Related works

### 2.1 GNN for PHM

The rise and development of GNNs have propelled the advancement of many research communities such as knowledge graphs [31] and social networks [32]. Due to the feature extraction ability of GNNs for unstructured data [33], scholars have introduced GNNs and their variants into the PHM of complex systems. For example, Chen et al. [34] introduced prior knowledge into diagnosis models and proposed a weighted GCN for circuit structure fault diagnosis. Zhao et al. [35] designed a multi-scale graph convolutional kernel for processing vibration signals to realize rotor-bearing fault diagnosis. Chen et al. [36] reviewed GNN-based methods for fault diagnosis from the perspectives of data processing, network architecture selection, experimental validation, and benchmark datasets. Nevertheless, studies that apply GNNs to various industrial images for defect detection are rare. Few attempts regarding GNNs as a complement to CNNs, while image feature extraction still adopts convolutional backbone networks [37, 38].

### 2.2 RPN in object detection

The RPN [39] refers to the region proposal network in the object detection algorithm. RPN plays a significant role in terms of generating predefined bounding boxes and performing preliminary defect classification and localization. For example, Tao et al. [40] introduced RPN into their detection model to compute proposals efficiently. Su et al. [41] proposed a novel RPN with channel attention and spatial attention to predict potential defect locations. Shih et al. [42] proposed a reduced RPN in object detection networks for reducing parameters and improving accuracy. In addition, scholars set the proposals with multiple scales in RPN due to the consideration of the size of defects [43, 44]. Since the initial bounding boxes generated in RPN are further refined to tightly surround the objects, the improvement of detection performance benefits from defect perception bounding boxes that exclude background interference is foreseeable.

## 2.3 Superpixels in various detection tasks

Superpixels provide a perceptual and progressive feature representation by grouping structurally or visually similar pixels in the images. Therefore, superpixel segmentation is widely used in various detection tasks to endow models with perception capabilities. For example, Zhang et al. [45] proposed a weighted coupling network by combining superpixel segmentation and a deep CNN for change detection of remote sensing images. Lv et al. [46] extracted superpixel features and fed them into the contractive autoencoder for change detection of SAR images. Tao et al. [47] improved the saliency detection performance of matrix factorization by sorting superpixels with spatial information. Zhou et al. [48] obtained the region of interest based on entropy rate superpixels and proposed a defect detection method utilizing attention strategy and wavelet transform. Thus, perceptive superpixel features not only bring performance in defect detection but also be integrated into existing models for a unified and efficient network pipeline.

## 3 The proposed method

Our work is an initial attempt to utilize GNN algorithms for intelligent defect detection of aeroengine blades. The are two challenges of applying GNN algorithms for intelligent defect detection of aeroengine blades. First of all, how to construct the graphs based on the images and how to conduct effectively aggregation and update of node features for feature extraction. To do that, we firstly divide the image into several patches and create graph based on these patches by viewing patches and pixel intensities as nodes and node features respectively, then we design a multi-stage GCN as backbone to extract hierarchical features. Therefore, it avoids high computation afford compared with the common graph construction methods due to a large number of pixels within the image. In addition, we design multi-head graph convolution operation like multi-head attention in Transformer to extract node features from different subspaces. The second challenge raises in two-stage defect detection pipeline, and how to combine extra information with feature extraction backbone for region proposal. Previous research works typically utilize features extracted by CNNs to generate bounding boxes, and the image processing in regular grids ignores the interaction between objects in an image. To this end, we jointly consider regular image features from feature extraction backbone and flexible superpixel features from superpixel segmentation branch. In this section, we will elaborate on the proposed SPGNN for intelligent defect detection, and the overall framework as shown in Fig. 1. The SPGNN consists of MSGCN for feature extraction and SPRPN for region proposal. The MSGCN extracts hierarchical defect features from the graph representation learning perspective, and the SPRPN provides perception properties by a superpixel segmentation on the graph features. As a result, the information transmission is always implemented at the graph level, and both benefit from each other in the whole network pipeline.

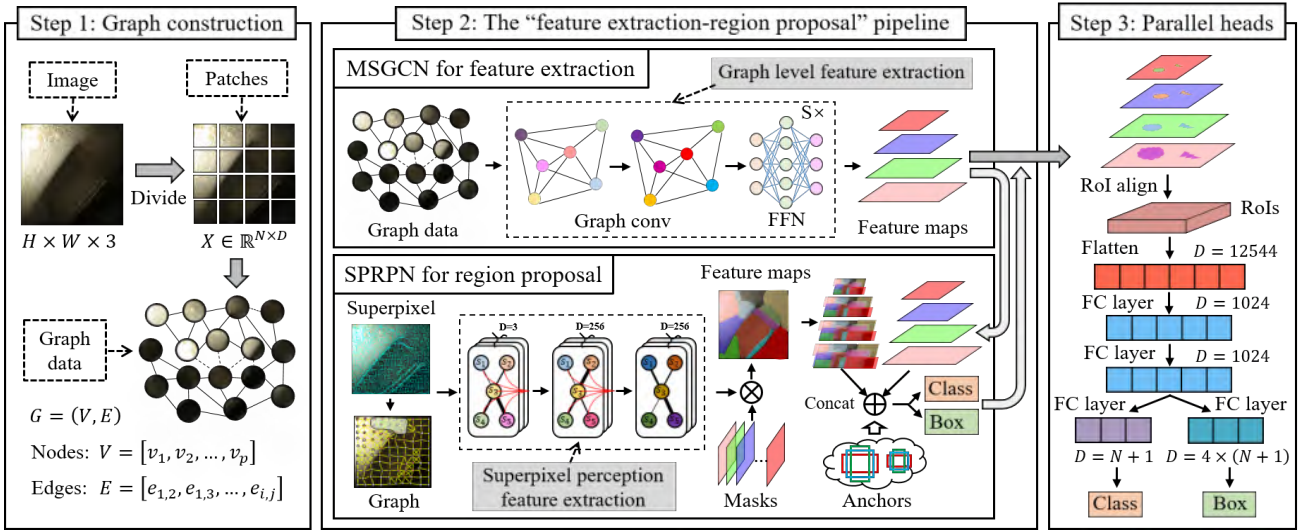

Fig. 1. The schematic diagram of the proposed SPGNN. First, borescope image with the size of $H \times W \times 3$ is divided image patches $X \in \mathbb{R}^{N \times d}$, then the graph is constructed by viewing each patch as node and the relation between each node and its neighbors as edge. Second, the graph data is sent into MSGCN, which consists of multi-stage graph convolutional layers and FFNs, for graph level feature extraction. The results of superpixel segmentation are processed into superpixel graph, then superpixel perception features are extracted by a two-layer graph convolutional network in SPRPN for region proposal. Final, the hierarchical features are used to classify and locate damages by two parallel heads.

## 3.1 Problem formulation

1) *Object Detection*: Blade defect detection can be considered as a realistic scenario of the general object detection task in computer vision. The purpose of this task is to find a detector that classifies and localizes objects (i.e., defects in our task) in images by training a large number of image-label pairs.

The formulations of the process are given by

$$\Theta = arg \min_{\Theta} \ell(y_i, \hat{y}_i) \ s.t. \ \hat{y}_i = Objector_{training}(x_i; \Theta) \quad (1)$$

$$z_i = Objector_{testing}(x_i; \Theta) \quad (2)$$

where $x_i$ represents the image data, $y_i$ is the corresponding annotated label, and $z_i$ denotes the inference results, i.e., defect category and coordinates. $Objector$ refers to the deep neural network with learnable parameters $\Theta$. $Objector_{training}$ denotes the defect detector which updates parameters through a large number of image-label pairs in training stage, and $Objector_{testing}$ denotes defect detector with trained (updated and fixed) parameters in testing stage, the model inputs defected blade images $x_i$ and outputs defect detection results $z_i$. In other words, $Objector_{testing}$ is the final updated and iterative results of defect detector $Objector_{training}$.

Specifically, assuming an annotated dataset $D = \{X, Y\}$ with a large number of image-label pairs, i.e., $X = \{x_1, x_2, \dots, x_N\}$ and $Y = \{y_1, y_2, \dots, y_N\}$, where $X$ and $Y$ denote the images and labels, respectively, and $N$ denotes the sample number of the dataset $D$. Going further, each $y_1 = \{l_1, l_2, \dots, l_K\}$ contains $K$ defined instances in an image, and the $i$-th object instance $l_i = [c_i, x_i, y_i, w_i, h_i]$, where $c_i$ denotes the defect category and $x_i, y_i, w_i, h_i$ denote the center coordinates, width, and height of the bounding box. In addition, we consider the defined defect categories as $C = \{1, 2, \dots, M\} \subset \mathbb{N}^+$, where $M$ denotes the number of defined categories.

2) *Graph Representation Learning*: The two key components of graph representation learning are graph construction and graph information aggregation (i.e., graph feature extraction). Considering a constructed graph $G = (V, E)$ with nodes $V = [v_1, v_2, \dots, v_p]$ and edges $E = [e_{1,2}, e_{1,3}, \dots, e_{i,j}]$. The node $v_i$ usually is created according to the number of people in social network, the number of sequences in vibration signal, or number of patches in the image, and the edge $e_{i,j}$ denotes the relation between the node $v_i$ and the node $v_j$. The graph information aggregation can be formulated as follows

$$f^l_{(i,j)} = \sum_{j \in n_i} \varphi^l_{agg}\left(f^l_j, f^l_{e_{(i,j)}}\right) \qquad (3)$$

$$f^{l+1}_i = \psi^l_{update}\left(f^l_j, f^l_{(i,j)}, W^l\right) \qquad (4)$$

where $f^l_{(i,j)}$ and $f^l_i$ denote the embedding of edge $e_{(i,j)}$ and the features of node $i$ at the $j$-th network layer, respectively. $\varphi(\cdot)$ and $\psi(\cdot)$ are the aggregation function and update function, respectively. $n_i$ denotes the set of neighborhoods of node $i$.

Therefore, a single layer of GNNs can be defined as follows

$$H^{l+1} = h(\hat{A} H^l W^l) \qquad (5)$$

where $H^l$ and $H^{l+1}$ denote the input and output of GNNs at the $j$-th layer, $\hat{A}$ represents a processed correlation matrix, $W^l$ denotes the learnable parameter matrix, and $h(\cdot)$ is the activation function.

## 3.2 MSGCN

The backbone, which is the feature extraction network, plays a significant role in various computer vision tasks. Two popular and widely used network architectures are CNNs and Transformers. However, they are hindered when extracting irregular and complex texture features. In particular, the limited receptive field of CNNs makes the network unable to capture long-range contextual information. While Transformers cannot model topological information. We summarize the differences between these three network architectures as shown in Fig. 2.

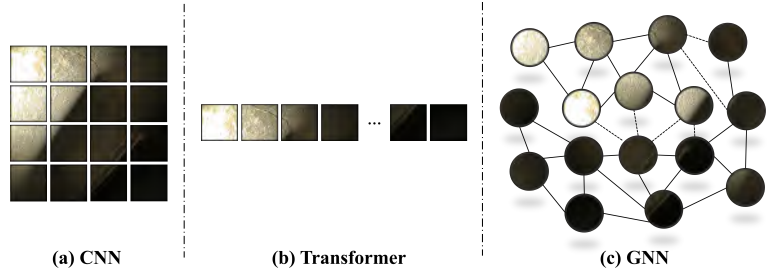

**(a) CNN**　　　**(b) Transformer**　　　**(c) GNN**

Fig. 2. The differences between CNNs, Transformers, and GNNs. (a) CNNs extract features in local regular grids by sliding convolutional operator over the entire image. (b) Transformers divide image data into a sequence of image patches, and view image feature extraction as long-range dependencies modeling. (c) GNNs construct a graph by viewing image patches and relations between different patches as nodes and edges, respectively. Thus, GNNs obtain non-local receptive field and adaptive feature extraction by conducting information transmission at graph level.

The first work that applies GNNs backbone for large-scale visual tasks was presented not long ago [49]. To obtain the input of SPGNN, we first process the initial image $I$ with the size of $H \times W \times 3$ by dividing it into $N$ patches with the feature dimension of $D$, i.e., we have a feature matrix $X \in \mathbb{R}^{N \times D}$. Then, we construct a graph $G = (V, E)$ for each image. Specifically, we treat each image patch as a node and obtain a set of nodes $V = [v_1, v_2, \ldots, v_N]$. While, we seek $K$ nearest neighbors for each node $v_i$ and crate a directed edge $e_{i,j}$ for $K$ nodes, i.e., $E_i = [e_{i,1}, e_{i,2}, \ldots, e_{i,K}]$, and $K = 9$ in our all experiments. Thus, MSGCN is naturally applied on a series of graph patches for defect feature extraction from a fresh graph representation learning perspective.

MSGCN consists of multi-stage GCN blocks, and graph convolutional layers and feed-forward networks (FFNs) make up each GCN block. Without loss of generality, we start with the input $X \in \mathbb{R}^{N \times D}$ of the MSGCN. According to the graph $\mathcal{G} = G(X)$ constructed above, graph convolutional layers can be applied on $\mathcal{G}$ for graph level feature extraction and information aggregation. This process is expressed as follows:

$$\mathcal{G}' = \mathcal{F}(\mathcal{G}, \mathcal{W}) \tag{6}$$

where $\mathcal{G}'$ denotes the updated graph, $\mathcal{F}$ denotes a combination of aggregation and update functions, and $\mathcal{W}$ denotes the learnable weights in $\mathcal{F}$. More specifically, max-relative graph convolution [50] is adopted in this paper, and it can be expressed as follows:

$$x'_i = \psi(x_i, \varphi(x_i, K(x_i), W_\varphi), W_\psi) \tag{7}$$

$$\begin{aligned}\varphi(\cdot) &= max(\{x_i - x_j | j \in K(x_i)\}) \\ \psi(\cdot) &= \varphi(\cdot)W_\psi\end{aligned} \tag{8}$$

where $\varphi(\cdot)$ and $\psi(\cdot)$ denote the aggregation and update functions, respectively, $W_\varphi$ and $W_\psi$ denote the learnable weights of aggregation and update functions, respectively. $K(x_i)$ denotes the $K$ nearest neighbors of the node $v_i$.

Compared with traditional GCN, the graph convolutional layers of GCN blocks in MSGCN introduce multi-head graph convolution operation like multi-head attention in Transformers. The node features with information aggregation are divided into $h$ heads to update node features in respective subspaces, and the information is concatenated to obtain final feature maps.

$$\begin{aligned}&head^1 W_\psi^1, head^2 W_\psi^2, \dots, head^h W_\psi^h \\ &head^h = \varphi(x_i, K(x_i), W_\varphi)^h\end{aligned} \tag{9}$$

After the above processing, the initial input $X \in \mathbb{R}^{N \times D}$ is processed as the output $X' \in \mathbb{R}^{N \times D'}$, and we define the graph processing process as $X' = GCN(X)$.

The issue of over-smoothing in GCNs will lead to indiscriminate node features and poor performance results for blade defect detection. To mitigate the negative effect of this phenomenon, FFN and GeLU are introduced into the MSGCN to enrich feature diversity and provide nonlinear attributes. In addition, the residual connection is applied in each GCN and FFN, as shown in Fig. 3. Therefore, each GCN block is formulated as follows:

$$Y = \sigma(GCN(XW_i))W_o + X \tag{10}$$

$$Z = \sigma(YW_i)W_o + Y \tag{11}$$

where $W_i$ and $W_o$ denote the weights of FFN, and $\sigma(\cdot)$ denotes the GeLU activation function.

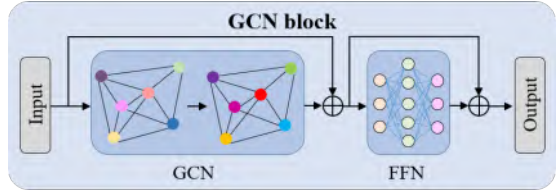

Fig. 3. The schematic diagram of GCN block. The GCN block consists of GCN for graph feature extraction and FFN for feature transformation. GeLU is the nonlinear activation function, and the residual connection is applied in each GCN and FFN.

To generate hierarchical features for multi-scale defect detection, the feature pyramid network is constructed by adding a downsample layer after each stage which includes several GCN blocks, as shown in Fig. 4. Firstly, the input with the size of $H \times W$ is processed through a convolutional layer to reduce the resolution of the image. Secondly, the constructed graph with $N$ nodes is fed into a multi-stage network architecture to obtain the feature pyramid with decreasing resolution and increasing feature dimension. Finally, the hierarchical features are used for RPN and downstream detection heads. The numbers of GCN blocks in the four stages are 2, 2, 6, and 2.

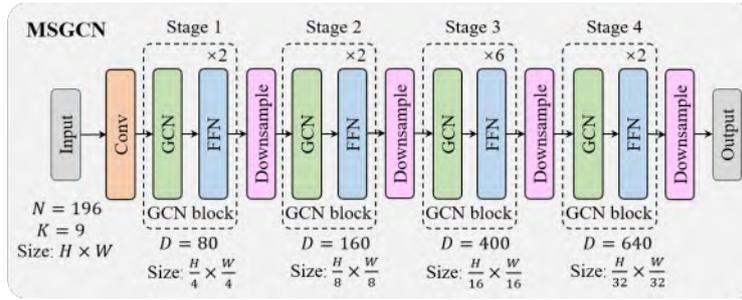

Fig. 4. The architecture diagram of MSGCN. The MSGCN takes graph representation feature as the input, and outputs the hierarchical feature pyramid. The size of feature map gradually decreases from $\frac{H}{4} \times \frac{W}{4}$ to $\frac{H}{8} \times \frac{W}{8}$, $\frac{H}{16} \times \frac{W}{16}$, and finally to $\frac{H}{32} \times \frac{W}{32}$. While, the dimension of the feature gradually increases from 80 to 160, 400, and finally to 640.

## 3.3 SPRPN

RPN plays an essential role in the two-stage object detection algorithm. Previous research works typically utilize features extracted by CNNs to generate bounding boxes, and the image processing in regular grids ignores the interaction between objects in an image, i.e., perception. To this end, we proposed the SPRPN for region proposal by jointly considering the regular CNN features in the image and flexible GNN features in superpixel segmentation.

As shown in Fig. 5, the superpixels $S = [s_1, s_2, ..., s_M]$ are extracted by applying SLIC which is a superpixel segmentation algorithm on the original image pixel $P = [p_1, p_2, ..., p_N]$ in SPRPN [51]. Firstly, we create node features for a graph $\mathcal{G} = \mathcal{G}(X)$ by viewing each superpixel $s_i$ as the node $v_i$ and calculating each superpixel intensity $F_i$ with

$$F_{v_i} = F_{s_i} = \frac{1}{L_i}\sum_{(x_i,y_i \in s_i)} F(x_i, y_i) \tag{12}$$

where $F_{v_i} = F_{s_i}$ denotes the node features, $L_i$ represents the number of pixels belonging to the superpixel $s_i$, and $F(x_i, x_j)$ denotes RGB feature of pixel $(x_i, x_j)$ in the original image. Therefore, the initial feature dimension is 3.

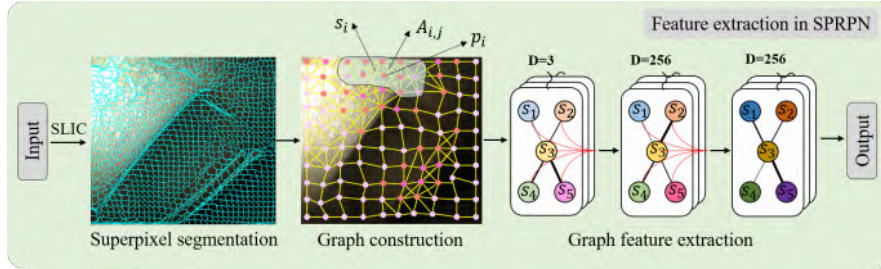

Fig. 5. The schematic diagram of feature extraction in SPRPN. SLIC algorithm is used to generate the result of superpixel segmentation. Then, superpixel graph is constructed as input to GCN for extracting superpixel perception features. The final feature dimension is 256, matching feature dimension from MSGCN. $p_i$ denotes a pixel in an image, $s_i$ denotes a superpixel, i.e., node, and $A_{i,j}$ denotes the adjacency matrix, i.e., edge.

Then, we conventionally crate edges between nodes in the form of an adjacency matrix. For the result of superpixel segmentation, the adjacency matrix $A_{i,j}$ is calculated between node $v_i$ and $v_j$ according to the coordinates of superpixels with

$$A_{i,j} = exp\left(-\frac{\|(x_i,y_i)-(x_j,y_j)\|^2}{\sigma^2_{(x,y)}}\right) \tag{13}$$

where $(x_i, y_i)$ and $(x_j, y_j)$ are the coordinates of a pair of superpixels $i$ and $j$. $\sigma^2_{(x,y)}$ denotes the average distance between $K$ neighbors and the node $v_i$. In our experiments, we adopt a Gaussian function to represent the scale factor, i.e., $\sigma^2_{(x,y)} = 0.1\pi$.

After having nodes and edges, we spontaneously design a two-layer GCN to extract superpixel perception features to endow the model with natural object-aware properties. Specifically, the two-layer GCN takes the constructed node features as input. The node features are aggregated and updated at graph level, and the feature dimension changed from 3 to 256, to match the dimension of the feature pyramid from MSGCN.

Finally, we delicately design a superpixel feature pyramid network for feature fusion in SPRPN, as shown in Fig. 6. Concretely, we append a mask $M_i$ for each graph superpixel feature $N \times F_i$. In this case, the features of each node will be assigned to the corresponding region of the original image via corresponding mask to obtain the recovered feature map with the size of $H \times W \times F_i$, i.e., $896 \times 896 \times 256$ in our experiments. Then, a feature pyramid network $\{S_2, S_3, S_4, S_5\}$ is constructed by applying $3\times 3$ convolution operators to the recovered feature map with different strides. In this way, the sizes of each level in superpixel feature pyramid network are consistent with those in image feature pyramid network $\{F_2, F_3, F_4, F_5\}$ for feature fusion. The superpixel feature pyramid from SPRPN and image feature pyramid from MSGCN are integrated by channel concatenation or element addition due to the consistency of feature map sizes. It is worth mentioning that we use additional $3\times 3$ convolutional operators to process the fused features $\{P_2, P_3, P_4, P_5\}$ to alleviate the aliasing effect of feature fusion. Consequently, the graph representation features from MSGCN and superpixel perception features are integrated for the region proposal in SPRPN.

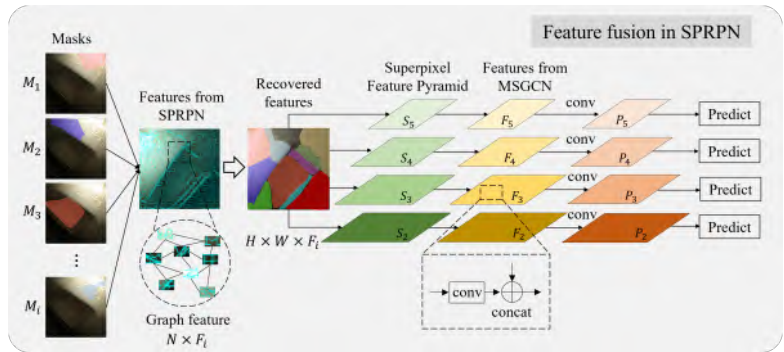

Fig. 6. The schematic diagram of feature fusion in SPRPN. According to the results of superpixel segmentation, we add a set of masks $M = [M_1, M_2, ..., M_i]$ for graph feature $N \times F_i$ from SPRPN, the size of feature maps is changed to the size of original image size $H \times W \times F_i$ for feature fusion.

## 3.4 Technical implementation

Based on open-source toolbox MMDetection, we can achieve model development by converting trained TORCH model to ONNX model for fast model inference. In addition, it is possible and feasible to apply these models into the existing lightweight hardware, such as NVIDIA Jetson AI inference platform[†]. Therefore, we can deploy our models to the platform that equipped with existing borescope inspection equipment to form an intelligent borescope inspection system.

In addition, due to the importance and uniqueness of aeroengine operation safety, it is necessary to pay attention to potential ethical and safety considerations. First of all, as we all know, model training requires a large number of aeroengine blade images. However, due to commercial copyright and confidentiality, it is necessary to coordinate between aeroengine maintenance departments, airlines, and aeroengine manufactures. In addition, the quality of data significantly affects defect detection model performance, and missed detection and false detection are both unacceptable in aeroengine inspection and maintenance. However, the defect annotation inevitably introduces human errors due to manual labeling process that relies on experience, and accidental mislabeling may lead to catastrophic results. Moreover, it is an issue worth exploring to iterate and update the detection mode in operational environment. The reason is that the model can only adapt to the changing operational environment if it continuously improves the ability to recognize unknown categories defects. In summary, within the scope of existing rules and ethical constraints, how to improve the accuracy and reliability of intelligent defect detection systems is of utmost importance.

---

[†] Link: https://www.nvidia.com/en-us/autonomous-machines/embedded-systems/jetson-orin/

## 4  Dataset construction

### 4.1 Blade datasets review

Unlike image datasets in other fields, due to commercial rights and copyright issues, there is still a lack of open source aero-engine blade datasets. In the industrial scenario of aero-engine blade damage detection, intelligent visual detection methods have just emerged, and currently there are only a few datasets. We briefly review existing blade datasets in this section. These datasets do not meet the needs of intelligent algorithm verification in borescope inspection for several reasons: (1) Images only have segmentation annotations, lacking classification labels and localization information. (2) Generally, infrared images and X-ray images are not acquired by digital cameras, but the results of digital imaging, which are not consistent with the scenarios of borescope inspection. (3) Some dataset contains a small number of images, which can easily lead to overfitting during training, and image resolution is often small due to cropping and resizing. We classify the existing blade image datasets into four categories as follows:

1) *X-ray image datasets* [52, 53]: these two X-ray image datasets include 2000 X-ray images [52] and 2137 X-ray images [53], respectively, which are acquired by an X-ray camera based on an aviation manufactory from the same research group. The first dataset has four types of defects: slag inclusion, gas cavity, crack, and cold shut, and the second dataset includes two more defect types: remainder, broken core, all of them are internal defects.

2) *Infrared image dataset* [54]: the dataset consists of 652 infrared images, including 600 non-defect images and 52 crack images. To verify the effectiveness of different methods, these infrared images are divided into 161 crack image patches and 7588 non-defect image patches for training and testing.

3) *Borescope images (single defect type) datasets* [55, 56]: these two borescope images datasets treat defect detection tasks as anomaly detection tasks. Some differences are that the dataset [55] which includes 420 borescope images with the size of $150 \times 150$ are divided into 4 types, and it is usually used to recognize defect types. A total of 1443 borescope images in another dataset [56] are collected from a local airline company in China, and it is used for defect segmentation tasks.

4) *Borescope images (multiple defect types) datasets* [57-59]: these three borescope image datasets are constructed based on field borescope inspection and inspection videos. Among them, two datasets that consist of 850 images [57] and 1916 images [58] respectively are processed via some data augmentation strategies, and they are used for defect detection tasks as the images are annotated with categories and localizations. The other dataset [59] involves 104 annotated images with five defect categories. These images are acquired from inspection videos and are annotated at pixel level, so it can be used for defect segmentation tasks.

## 4.2 Blade datasets construction

To verify the effectiveness of the proposed method, we elaborately construct a simulated blade dataset. Real blades are processed through 3D scanning, 3D drawing, and 3D printing to generate simulated blades. Various damages are prefabricated via laser machining technology, and blade images token by digital camera are annotated with classification label and localization information.

Specifically, the simulated blade dataset (SB dataset) with 3000 defect blade images and 5 defect modes is constructed by solid modeling, 3D printing, laser processing, photograph, and data annotation, as shown in Fig. 7. Firstly, we use 3D modeling software to generate a solid model of the blade based on the shape and curve of the real aero-engine blade. Secondly, we use 3D printing and laser machining for blade forming and defect prefabrication, respectively. Then, we use a digital camera to take pictures of the defective blade. Finally, we annotate the images using an image annotation tool called Labelme, including the defect types and locations. Specifically, there are five defect types in this dataset, including cracks, burned, broken, nick, and overheated. The size of image is $6000 \times 4000$, and the real size of the defect is marked in Fig. 8. Both single-defect and multi-defect images exist simultaneously, and the number of each type is shown in Table II.

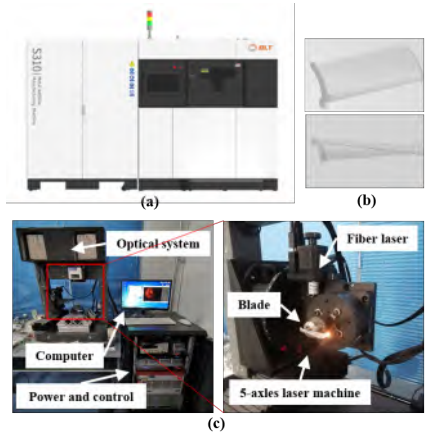

Fig. 7. The schematic diagram of simulated blade machining. (a) BLT-S310 3D printer. (b) The solid model of the blade. (c) Laser machining for prefabricating damages.

Table II. Statistical results of SB and AL datasets

| Category | Amount | Category | Amount |
|---|---|---|---|
| Broken | 250 | Non conducting | 390 |
| Burned | 250 | Abrasion mark | 128 |
| Crack | 750 | Corner leakage | 346 |
| Nick | 250 | Orange peel | 173 |
| Overheated | 250 | Under screen | 538 |
| Broken + Burned | 150 | Splashiags | 86 |
| Broken + Overheated | 100 | Paint bubble | 82 |
| Crack + Broken | 250 | Pit | 407 |
| Crack + Burned | 250 | Chafed | 365 |
| Crack + Crack | 250 | Dirty spot | 261 |
| Crack + Overheated | 250 | Multi-defects | 229 |
| SB dataset: 3000 | | AL dataset: 3005 | |

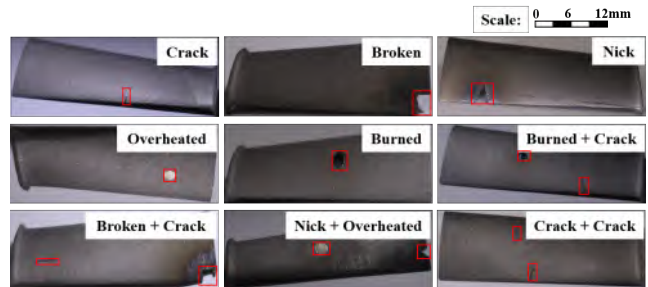

Fig. 8. The instances of simulated blade dataset.

In addition, we evaluate the proposed method on a public dataset with 3005 aluminum defect images. The aluminum dataset (AL dataset) consists of 3005 aluminum defect images with 10 defect types. The dataset is available at https://tianchi.aliyun.com/competition/entrance/231682/infomation on the Alibaba TIANCHI platform. As shown in Fig. 9, the sizes of different defects vary dramatically, the shapes of different defects are diverse, and the divergence of different defect is imperceptible. In addition, the class imbalance problem also exists in this dataset. For example, there are 390 "non conducting" defect images, and only 82 "paint bubble" defect images. The above characteristics bring great challenges to defect detection.

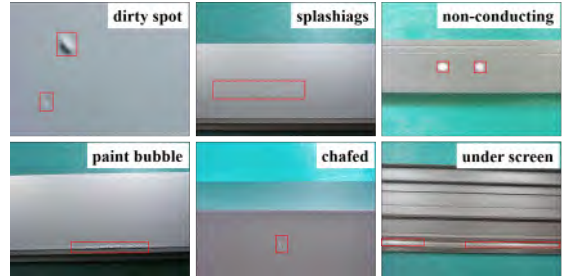

Fig. 9. The instances of aluminum dataset.

## 5 Experimental study

In this section, we carry out extensive experiments to evaluate the performance of the proposed SPGNN on two damage detection datasets, including a homemade SB dataset and a public AL dataset. The number of images and the types of defects are shown in Table II. The purposes of these experiments are as follows: 1) Demonstrate the effectiveness and superior performance of the proposed SPGNN for intelligent defect detection. 2) Present the excellent modeling ability of SPGNN for irregular and complex textures via visualization results of different methods. 3) Analyze the parameter sensitivity of the number of nodes neighbors in MSGCN and different feature fusion strategies in SPRPN. 4) Discuss the effectiveness of MSGCN and SPRPN by ablation experiments.

### 5.1 Experimental setup

*1) Comparison Methods*: We compare the proposed SPGNN with CNN-based methods, such as Faster R-CNN [39], PAFPN [60], Libra R-CNN [61], and Sparse R-CNN [62]. An illustrious Transformer-based method named Swin [63] is also used for performance comparison. Two industrial defect detection methods for blades [12] and fasteners [27] are also involved in our comparison experiments. The details of the comparison methods are as follows:

*a) Faster R-CNN*: Faster R-CNN [39] is the pioneering work that achieves end-to-end object detection. As a representative work of a two-stage object detection algorithm, Faster R-CNN generates initial bounding boxes by introducing RPN, improving the efficiency of the whole network.

*b) PAFPN*: PAFPN [60] rethinks the information transmission path in feature pyramid network and builds a path

aggregation network by appending a top-down feature augmentation module to the feature pyramid network.

*c) Libra R-CNN*: Libra R-CNN [61] aims to alleviate the imbalanced learning in object detection by introducing balanced IoU, balanced pyramid, and balanced L1 from sample level, feature level, and objective level, respectively.

*d) Sparse R-CNN*: Sparse R-CNN [62] is the novel CNN-based object detection algorithm. This work presents a sparse scheme by replacing RPN with a s parse set of learnable proposal boxes, forming a refreshing paradigm.

*g) Swin*: Swin [63] is a versatile backbone for various vision tasks. This work builds a hierarchical feature extraction network by merging neighboring patches. Especially, shifted window strategy is introduced into vanilla Transformer to obtain linear computation complexity w.r.t. image size.

*2) Implementation Details[‡]*: We use mAP[§], a common metric for object detection, to evaluate the performance of different models in our experiments. Here, we present a schematic diagram of this metric, as shown in Fig. 10. Also, you can find more details in the URL given in the footnote. In SB dataset, we use 2700 damaged blade images to train our model, and evaluation results are reported on the remained 300 images. In AL dataset, 2704 images are used to train our model, and 301 images are evaluated to compare the performances of different methods. Unless otherwise noted, the same data division were used in all our experiments. For all methods, we resize the input image to 896×896 for a fair comparison because of considering the significant impact of input image size. Due to the approximately same model capacities and parameters, we utilize the ImageNet pre-trained models, i.e., ResNet-50, Swin-T, and ViG-S, for CNN-based methods, Swin, and SPGNN, respectively in our experiments. We set the training schedule to 12 epochs for all experiments. Following the default settings of MMDetection, we adopt SGD as the optimizer for CNN-based methods and SPGNN and set the momentum and weight decay to 0.9 and 0.0001, respectively. While we adopt Adam as the optimizer for Swin. All experiments are implemented on MMDetection toolbox and benchmark.

---

[‡] The method [27] consists of three stages for fastener defect classification and localization in a coarse-to-fine manner, and we consider the stage of defect localization and compare it with our method.

[§] The detailed explanation and evaluation code of the mAP are available at https://cocodataset.org/#detection-eval.

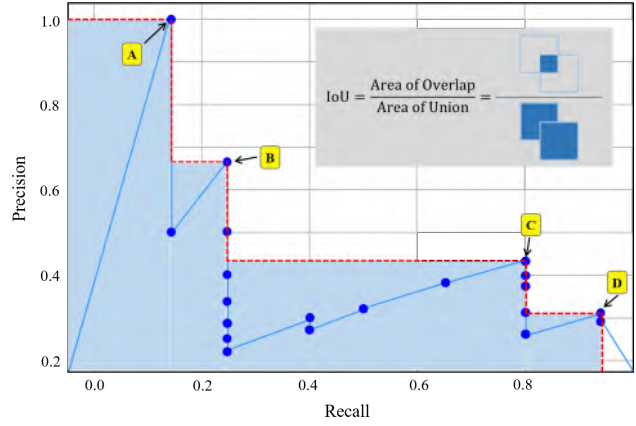

Fig. 10. The illustration of mAP and IoU. Firstly, we use IoU to evaluate whether a detection bounding box is a true positive. Then we calculate the precision/recall curve according to this criterion. Finally, we obtain the AP by computing the area under curve.

## 5.2 Defect detection performance

To evaluate the performances of different methods, we carry out the comparison experiments on SB dataset and AL dataset, the detection results of different methods as shown in Table III. Compared with other methods, the proposed method achieves the best performance with 56.4 mAP and 81.0 $AP_{50}$ on SB dataset. Besides, the proposed SPGNN obtains superior results with 34.5 mAP and 50.4 $AP_{50}$ on AL dataset. In general, the proposed SPGNN obtains consistent performance gains across different metrics. In addition, we also analyze the detection performance of different methods for different defect categories. The mAP results are shown in Table IV and Table V for SB dataset and AL dataset, respectively. According to the mAP results, we observe that the proposed SPGNN obtains excellent results on most defect categories. The anomalous phenomenon occurs on the class "pit", where sparse R-CNN obtains performance far exceeding other methods, the possible reason is that this method avoids the design of region proposals and NMS processing. All methods perform poorly on categories "abrasion mark" and "splashiags". In general, the proposed SPGNN achieves the best overall performance.

Table III. The experimental results of different methods on SB and AL datasets

| Methods | SB dataset | | | | | | | |
|---|---|---|---|---|---|---|---|---|
| | Backbones | Param (M) | mAP | $AP_{50}$ | $AP_{75}$ | mAR | $AP_M$ | $AP_L$ |
| Faster R-CNN [39] | ResNet-50 | 41.17 | 38.8 | 70.6 | 35.9 | 44.8 | 21.3 | 47.4 |
| PAFPN [60] | ResNet-50 | 41.43 | 48.4 | 77.4 | 52.3 | 52.6 | 31.2 | 56.5 |
| Libra R-CNN [61] | ResNet-50 | 41.43 | 52.2 | 77.1 | 58.7 | 56.5 | 35.7 | 60.9 |
| Sparse R-CNN [62] | ResNet-50 | 105.96 | 41.3 | 79.2 | 36.0 | 57.0 | 37.2 | 41.5 |
| Swin-based R-CNN | Swin-T [63] | 47.42 | 48.1 | 77.8 | 54.0 | 52.9 | 32.7 | 54.5 |
| Shang et al. [12] | ResNet-50 | 44.94 | 38.7 | 66.9 | 39.5 | 42.7 | 18.9 | 49.9 |
| Chen et al. [27] | Darknet-53 | 61.55 | 5.1 | 23.0 | 0.5 | 13.5 | 2.8 | 7.3 |
| SPGNN (ours) | ViG-S [49] | 46.56 | **56.4** | **81.0** | **65.9** | **61.1** | **42.1** | **61.8** |
| Methods | AL dataset | | | | | | | |
| | Backbones | Param (M) | mAP | $AP_{50}$ | $AP_{75}$ | mAR | $AP_M$ | $AP_L$ |
| Faster R-CNN [39] | ResNet-50 | 41.17 | 29.2 | 45.3 | 31.9 | 36.7 | 0.5 | 30.3 |
| PAFPN [60] | ResNet-50 | 41.43 | 30.4 | 47.2 | 32.3 | 39.0 | 0.7 | 31.5 |
| Libra R-CNN [61] | ResNet-50 | 41.43 | 28.3 | 44.8 | 28.7 | 36.0 | 0.7 | 29.3 |
| Sparse R-CNN [62] | ResNet-50 | 105.96 | 32.0 | 48.8 | 35.2 | 40.8 | 0.1 | 33.0 |
| Swin-based R-CNN | Swin-T [63] | 47.42 | 27.5 | 45.7 | 28.4 | 37.6 | 1.7 | 28.5 |
| Shang et al. [12] | ResNet-50 | 44.94 | 33.7 | 46.9 | 36.5 | 38.6 | 1.0 | 34.6 |
| Chen et al. [27] | Darknet-53 | 61.55 | 12.6 | 35.2 | 6.9 | 22.1 | 0.3 | 14.3 |
| SPGNN (ours) | ViG-S [49] | 46.56 | **34.5** | **50.4** | **38.0** | **42.6** | **4.3** | **50.4** |

Table IV. The mAP results of different damage types on SB dataset

| Types | Faster R-CNN | PAFPN | Libra R-CNN | Sparse R-CNN | Swin-based R-CNN | Shang et al. [12] | Chen et al. [27] | SPGNN (ours) |
|---|---|---|---|---|---|---|---|---|
| Broken | 32.6 | 43.4 | 46.3 | 21.9 | 40.7 | 29.0 | 0.9 | **45.9** |
| Burned | 47.8 | 60.1 | 64.3 | 65.9 | 59.1 | 52.6 | 13.2 | **65.9** |
| Crack | 15.6 | 22.2 | 27.4 | 23.4 | 22.3 | 11.9 | 0.7 | **32.0** |
| Nick | 46.1 | 51.9 | 54.8 | 40.0 | 52.6 | 40.9 | 4.3 | **62.4** |
| Overheated | 51.8 | 64.6 | 68.3 | 55.2 | 65.8 | 59.1 | 6.6 | **71.7** |

Table V. The mAP results of different damage types on AL dataset

| Types | Faster R-CNN | PAFPN | Libra R-CNN | Sparse R-CNN | Swin-based R-CNN | Shang et al. [12] | Chen et al. [27] | SPGNN (ours) |
|---|---|---|---|---|---|---|---|---|
| Nc | 47.2 | 45.4 | 47.7 | 39.6 | 44.3 | 53.8 | 4.1 | **54.7** |
| Am | 2.2 | **4.7** | 3.1 | 0.5 | 3.1 | 3.5 | 0.8 | 1.3 |
| Cl | 57.1 | 58.6 | 49.5 | 63.3 | 40.4 | 38.1 | 15.8 | **69.4** |
| Op | 40.8 | 42.7 | 44.4 | 44.2 | 41.7 | **54.8** | 12.2 | 46.9 |
| Us | 38.6 | 45.4 | 37.3 | 43.0 | 40.2 | 51.4 | 18.0 | **57.0** |
| Splashiags | 3.9 | 3.9 | 1.5 | 0.0 | 7.9 | **20.7** | 0.0 | 1.3 |
| Pb | 0.0 | 0.2 | 0.3 | 0.1 | 0.4 | 0.0 | 1.5 | **3.8** |
| Pit | 7.7 | 6.9 | 2.9 | **30.5** | 4.5 | 7.1 | 16.8 | 8.0 |
| Chafed | 81.7 | 82.6 | 82.0 | 86.1 | 82.5 | **92.0** | 48.2 | 87.8 |
| Dirty spot | 13.3 | 13.9 | 14.2 | 12.3 | 10.0 | **15.1** | 8.4 | **15.1** |

where Nc denotes non conducting, Am denotes abrasion mark, Cl denotes corner leakage, Op denotes orange peel, Us denotes under screen, Pb denotes paint bubble.

The reason for this phenomenon is that the MSGCN adaptively extracts de fect features because of flexible receptive field. Another important factor is that the proposed SPRPN generates bounding boxes with perceptual properties by integrating image features and superpixel features. The generated region proposals better focus the defect regions, which is beneficial to the improvement of detection performance. To verify our insights, we conduct ablation experiments to demonstrate the effects of MSGCN and SPRPN, the detailed experimental results are shown in Section VI.

## 5.3 Visualization results

To visually compare the performance of different methods, we visualize the detection results. The bounding boxes with categories and confidences are annotated on the images. As shown in Fig. 11 and Fig. 12, compared with other methods, the proposed SPGNN accurately localizes the defects with higher classification confidence. For example, many false detection boxes and missed detections appear in the detection results of other methods, and overlapping detection boxes also appear in the visualization results. Furthermore, these methods are not very confident and do not tightly wrap the contours of the defect. While our method achieves the best visual results, the highest confidence score, and the most accurate localization. The reason for the phenomenon is that SPRPN integrates the image features and superpixel features to generate the bounding boxes with perceptual capability and accurate localization for downstream tasks. Compared with vanilla RPN, the proposed SPRPN generates perceptual bounding boxes, further improving detection performance. In, addition. we empirically have the conclusion that a good feature extraction network will accurately focus on defect regions, and this principle can be visually reflected in the detection results. In this regard, feature extraction from the GNNs perspective in MSGCN is a promising backbone network suitable for the detection of irregular and complex defects. In addition, we test some real borescope images mentioned in Table I to clearly depict various damages to enhance the experimental results, the visualization results are shown in Fig. 13. One can see that the proposed SPGNN shows desired performance on real defected blade images.

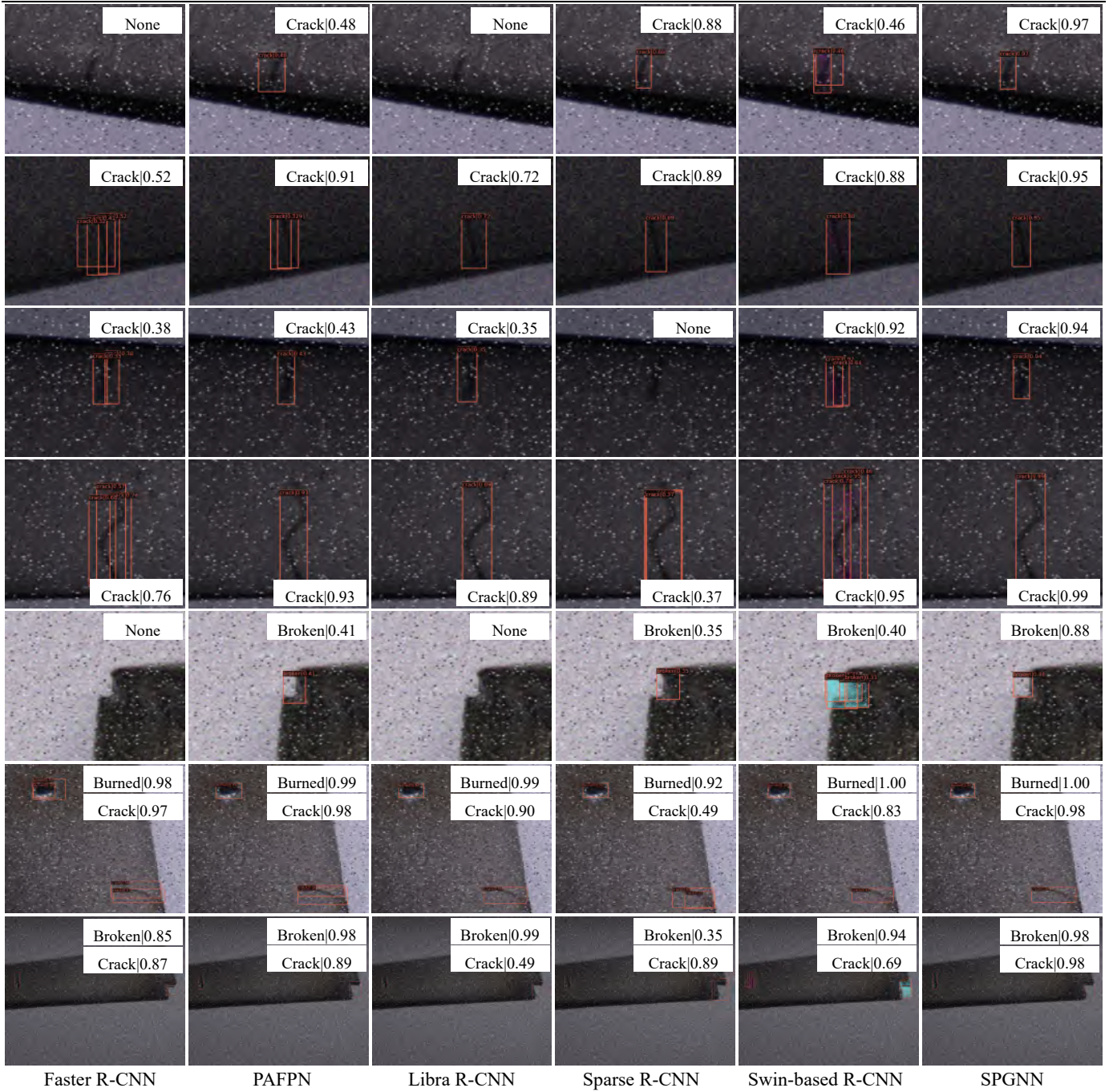

Fig. 11. The visualization results of different methods on SB dataset.

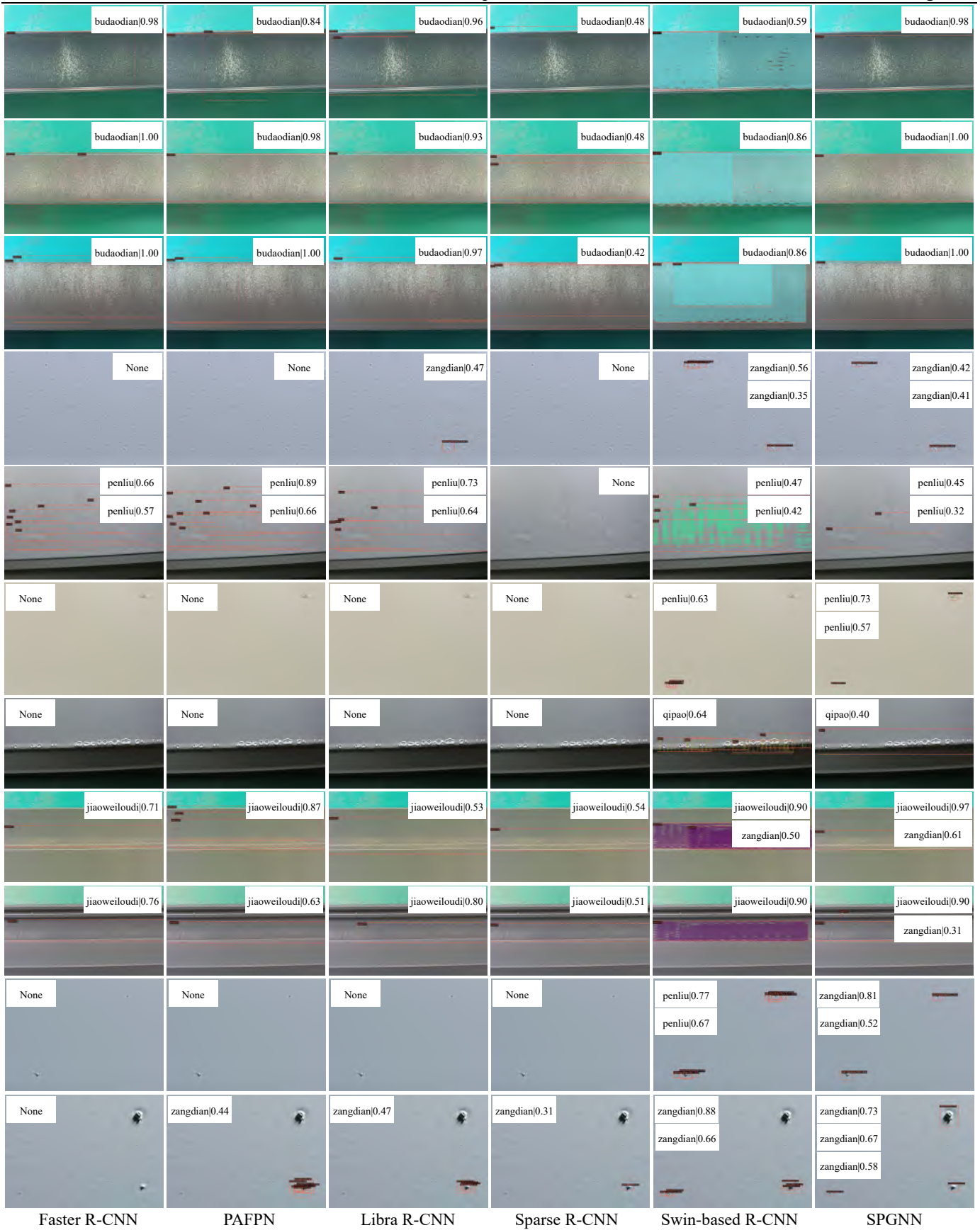

Fig. 12. The visualization results of different methods on AL dataset.

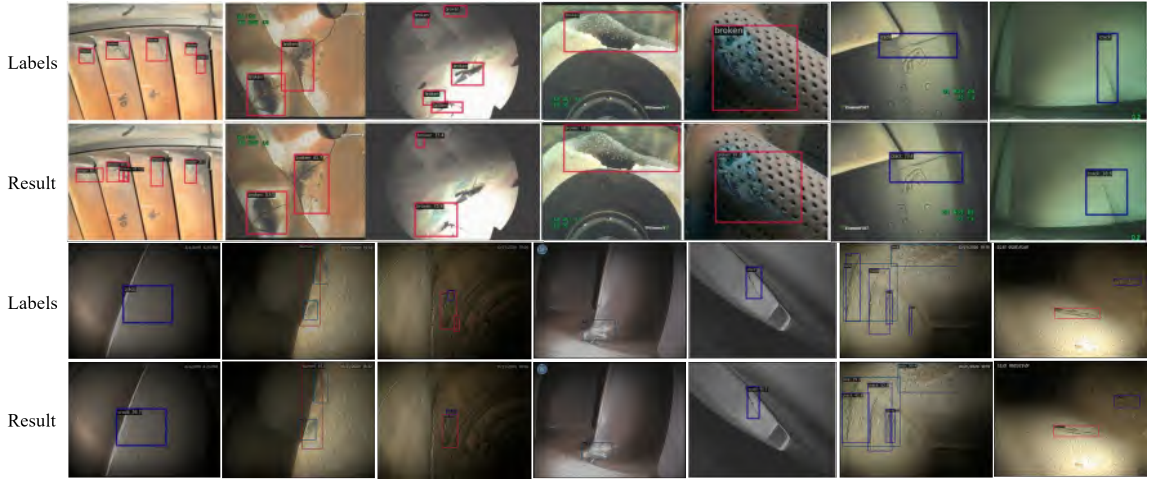

Fig. 13. The visualization results on some real borescope images.

## 5.4 Parameter sensitivity analysis

*1) Nearest Neighbors in MSGCN*: In MSGCN, the constructed graphs are processed through the multi-stage GCN. The number of nearest neighbors of nodes affects the size of the receptive field, which affects feature extraction and detection performance. The experimental results are shown in Table VI and Table VII. It is observed that 9 is an optimal choice for the number of nearest neighbors on both the SB dataset and the AL dataset, either 6 or 12 achieved suboptimal performance (78.9 $AP_{50}$ for k=6, 81.0 $AP_{50}$ for k=9, and 80.1 $AP_{50}$ for k=12 on SB dataset. 48.4 $AP_{50}$ for k=6, 50.4 $AP_{50}$ for k=9, and 48.2 $AP_{50}$ for k=12 on SB dataset). We analyze that small values lead to too little aggregated information, which is not conducive to feature extraction, while large values will introduce more noise, affecting the feature extraction and detection performance.

Table VI. The results of different super-parameters in MSGCN

| Neighbors K | SB dataset | | | | | |
| --- | --- | --- | --- | --- | --- | --- |
| | mAP | $AP_{50}$ | $AP_{75}$ | mAR | $AP_M$ | $AP_L$ |
| 6 | 54.5 | 78.9 | 61.5 | 58.7 | 40.1 | 59.2 |
| 9 | 56.4 | 81.0 | 65.9 | 61.1 | 42.1 | 61.8 |
| 12 | 54.1 | 80.1 | 62.3 | 58.3 | 38.6 | 60.1 |

Table VII. The results of different super-parameters in MSGCN

| Neighbors K | AL dataset | | | | | |
| --- | --- | --- | --- | --- | --- | --- |
| | mAP | $AP_{50}$ | $AP_{75}$ | mAR | $AP_M$ | $AP_L$ |
| 6 | 33.2 | 48.4 | 35.9 | 41.3 | 3.0 | 34.5 |
| 9 | 34.5 | 50.4 | 38.0 | 42.6 | 4.3 | 35.4 |
| 12 | 33.7 | 48.2 | 37.9 | 40.1 | 2.0 | 35.0 |

*2) Fusion Strategies in SPRPN*: In SPRPN, the image features and superpixel features are fused to obtain richer feature representations for region proposal. Therefore, we explore the impact of different fusion strategies on detection performance.

Specifically, we carry out the comparative experiments by adopting different feature fusion strategies in SPRPN, i.e., element addition and channel concatenation. The experimental results are shown in Table VIII and Table IX. It is observed that the channel concatenation strategy is the better choice for SPRPN (81.0 $AP_{50}$ v.s. 79.5 $AP_{50}$ on SB dataset and 50.4 $AP_{50}$ *v.s.* 49.0 $AP_{50}$ on AL dataset). The possible reason is that the channel concatenation of superpixel features and image features can effectively utilize features with different attributes and provide richer graph representation features.

Table VIII. The results of different fusion strategies in SPRPN

| Fusion strategy | SB dataset | | | | | |
| --- | --- | --- | --- | --- | --- | --- |
| | mAP | $AP_{50}$ | $AP_{75}$ | mAR | $AP_M$ | $AP_L$ |
| add | 54.3 | 79.5 | 63.7 | 58.4 | 40.4 | 58.4 |
| concat | 56.4 | 81.0 | 65.9 | 61.1 | 42.1 | 61.8 |

where "add" denotes addition of feature elements, and "concat" denotes concatenation of feature channels.

Table IX. The results of different fusion strategies in SPRPN

| Fusion strategy | AL dataset | | | | | |
| --- | --- | --- | --- | --- | --- | --- |
| | mAP | $AP_{50}$ | $AP_{75}$ | mAR | $AP_M$ | $AP_L$ |
| add | 32.5 | 49.0 | 36.9 | 39.0 | 2.7 | 33.3 |
| concat | 34.5 | 50.4 | 38.0 | 42.6 | 4.3 | 35.4 |

where "add" denotes addition of feature elements, and "concat" denotes concatenation of feature channels.

# 6 Ablation experiment

In this section, we conduct ablation experiments to verify the effects of MSGCN and SPRPN. Specifically, we replace the original backbone with MSGCN in Faster R-CNN and observe the performance gains from MSGCN. For the proposed SPRPN, we apply the SPRPN to different methods to explore the impact of superpixel perception features on performance.

## 6.1 The effects of MSGCN

Efficient feature extraction plays a significant role in defect detection, we explore the feasibility of graph representation learning for feature extraction in our work. To this end, we conduct ablation experiments by replacing the CNN-based backbone in Faster R-CNN with MSGCN. As shown in Table X and Table XI, it is observed that MSGCN brings significant improvement with 8.6 $AP_{50}$ on SB dataset and gains 1.8 $AP_{50}$ benefits on AL dataset. In addition, it can observe that the gain brought by the coexistence of MSGCN and SPRPN is the largest, which is greater than the performance improvement brought by the separate existence. The reason for this phenomenon is that the "feature extraction-region proposal" pipeline always implements feature extraction and information transmission at the graph level. Therefore, the pure graph representation learning pipeline does not consider the barriers to information transmission between different network architectures, i.e., MSGCN and SPRPN mutually benefit from each other according to the experimental results.

Table X. The ablation experiments on SB dataset

| Method | MSGCN | SPRPN | mAP | $AP_{50}$ | $AP_{75}$ | mAR |
|---|---|---|---|---|---|---|
| Faster R-CNN | | | 38.8 | 70.6 | 35.9 | 44.8 |
| | √ | | 52.3 | 79.2 | 58.3 | 56.1 |
| | | √ | 45.9 | 77.8 | 49.1 | 50.8 |
| | √ | √ | **56.4** | **81.0** | **65.9** | **61.1** |

Table XI. The ablation experiments on AL dataset

| Method | MSGCN | SPRPN | mAP | $AP_{50}$ | $AP_{75}$ | mAR |
|---|---|---|---|---|---|---|
| Faster R-CNN | | | 29.2 | 45.3 | 31.9 | 36.7 |
| | √ | | 31.3 | 47.1 | 35.1 | 40.2 |
| | | √ | 31.8 | 46.8 | 35.7 | 38.5 |
| | √ | √ | **34.5** | **50.4** | **38.0** | **42.6** |

## 6.2 The effects of SPRPN

Region proposal is a crucial link in the two-stage object detection algorithm, and the quality of the initial bounding box significantly affects the detection performance. In our work, we carry out ablation experiments to investigate the effects of the proposed SPRPN by replacing vanilla RPN with SPRPN. As shown in Table XII and Table XIII, after replacing the vanilla RPN with SPRPN, continuous performance gains are obtained on both SB dataset and AL dataset. We can conclude that utilizing superpixel features and image features to generate perceptual bounding boxes is a promising insight in SPRPN. Superpixel segmentation features have natural object-aware properties due to considering the similarity of features such as color, brightness, and texture. Due to taking account of perceptual similarity relations, the superpixel perception bounding boxes are beneficial for improving detection performance.

Table XII. The ablation experiments on SB dataset

| Method | w/o SPRPN | mAP | $AP_{50}$ | $AP_{75}$ | mAR |
|---|---|---|---|---|---|
| Faster R-CNN |  | 38.8 | 70.6 | 35.9 | 44.8 |
|  | √ | 45.9 | 77.8 | 49.1 | 50.8 |
| PAFPN |  | 48.4 | 77.4 | 52.3 | 52.6 |
|  | √ | 50.0 | 77.7 | 58.1 | 54.1 |
| Libra R-CNN |  | 52.2 | 77.1 | 58.7 | 56.5 |
|  | √ | 54.9 | 78.4 | 64.6 | 58.2 |
| Swin-based R-CNN |  | 48.1 | 77.8 | 54.0 | 52.9 |
|  | √ | 50.2 | 81.7 | 56.6 | 54.6 |

Table XIII. The ablation experiments on AL dataset

| Method | w/o SPRPN | mAP | $AP_{50}$ | $AP_{75}$ | mAR |
|---|---|---|---|---|---|
| Faster R-CNN |  | 29.2 | 45.3 | 31.9 | 36.7 |
|  | √ | 31.8 | 46.8 | 35.7 | 38.5 |
| PAFPN |  | 30.4 | 47.2 | 32.3 | 39.0 |
|  | √ | 33.3 | 49.4 | 37.8 | 38.6 |
| Libra R-CNN |  | 28.3 | 44.8 | 28.7 | 36.0 |
|  | √ | 31.2 | 48.8 | 31.8 | 38.5 |
| Swin-based R-CNN |  | 27.5 | 45.7 | 28.4 | 37.6 |
|  | √ | 29.5 | 47.3 | 31.3 | 40.0 |

# 7 Conclusion

In this paper, we propose a novel method dubbed SPGNN for intelligent defect detection. The "feature extraction-region proposal" pipeline with graph representation learning in SPGNN is analyzed and discussed comprehensively and thoroughly. The experimental results on SB and AL datasets show that the proposed SPGNN has superior performance compared with state-of-the-art methods. We summarize the key insights in our work as follows:

1) *Feature extraction from GNNs perspective*: We implement defect detection from the perspective of graph representation learning by formulating local grid processing and defect feature extraction as graph structure construction and graph information aggregation, respectively. The extracted features have richer contextual information due to the larger receptive field. It also overcomes the inadequacy of CNNs to process images in local regular grids, which is beneficial for the detection of irregular and complex defects.

2) *Region proposal via superpixel perception*: Superpixel segmentation features have natural object-aware properties due to considering the similarity of features such as color, brightness, and texture. Our insight is utilizing superpixel features and image features to generate perceptual bounding boxes in RPN. Due to taking account of perceptual similarity relations, the superpixel perception bounding boxes are beneficial for improving detection performance.

3) *Information transmission at graph level*: In SPGNN, the "feature extraction-region proposal" pipeline always implements feature extraction and information transmission at the graph level. Therefore, the pure graph representation learning pipeline does not consider the barriers to information transmission between different network architectures.

Although the proposed method shows good performance on surface defect detection of aero-engine blade, there are two main limitations of our method. We summarize the limitations of the proposed method and discuss future steps.

1) Unified defect detection pipeline with data-driven and adaptive superpixel segmentation: Superpixel generation is time-consuming because we need to generate the corresponding superpixel results via SLIC algorithm in advance. Besides, the SLIC algorithm is not a learning-based algorithm, it is separate from the SPGNN defect detection network pipeline. We have to add a set of masks for graph feature to achieve the goal of fusing image features and superpixel features to generate bounding boxes with perceptual properties in SPRPN. In the future, we will focus on data-driven and adaptive superpixel generation methods through a learnable network pipeline.

2) Continuous learning-based defect detection methods for unknown types recognition: Our method is currently trained on a large amount of labeled data, and data collection and labeling are labor intensive. In addition, our method is not effective at detecting untrained defect types, the problem limits the application of intelligent defect detection methods in realistic scenario. In the future, we will explore towards open-world solutions based on continuous learning and incremental learning theories to address the problem of detecting unknown types.

# Reference

[1] H. Shang, J. Wu, C. Sun, J. Liu, X. Chen, R. Yan, " Global Prior Transformer Network in Intelligent Borescope Inspection for Surface Damage Detection of Aeroengine Blade," IEEE T. Ind. Inform. 2023.

[2] Y. Xiao. H. Shao, M. Feng, T. Han, J. Wan, B. Liu, "Towards trustworthy rotating machinery fault diagnosis via attention uncertainty in Transformer," Journal of Manufacturing Systems, 70, 186-201. 2023.

[3] D. Li, J. Chen, R. Huang, Z. Chen, W. Li, "Sensor-aware CapsNet: Towards trustworthy multisensory fusion for remaining useful life prediction," Journal of Manufacturing Systems, 72, 26-37. 2024.

[4] Y. Liu, H. Jiang, R. Yao, H. Zhu, "Interpretable data-augmented adversarial variational autoencoder with sequential attention for imbalanced fault diagnosis," Journal of Manufacturing Systems, 71, 342-359. 2023.

[4] H. Shang, C. Sun, J. Liu, X. Chen, R. Yan, "Defect-aware transformer network for intelligent visual surface defect detection," Advanced Engineering Informatics, 55, 101882. 2023.

[6] Y. Gao, X. Li,. X. V. Wang, L. Wang, L. Gao, A Review on Recent Advances in Vision-based Defect Recognition towards Industrial Intelligence, Journal of Manufacturing Systems, 62, 753-766. 2022.

[7] D. Wang, Y. Chen, C. Shen, J. Zhong, Z. Peng, and C. Li, "Fully interpretable neural network for locating resonance frequency bands for machine condition monitoring," Mech. Syst. Signal Pr., vol. 168, p. 108673. 2022.

[8] X. Li, H. Shao, S. Lu, J. Xiang, and B. Cai, "Highly Efficient Fault Diagnosis of Rotating Machinery Under Time-Varying Speeds Using LSISMM and Small Infrared Thermal Images," IEEE Transactions on Systems, Man, and Cybernetics: Systems. 2022.

[9] C. Wang, N. Lu, Y. Cheng, and B. Jiang, "A data-driven aero-engine degradation prognostic strategy," IEEE T. Cybernetics, vol. 51, no. 3, pp. 1531-1541. 2019.

[10] K. Hu, Y. Cheng, J. Wu, H. Zhu, and X. Shao, "Deep Bidirectional Recurrent Neural Networks Ensemble for Remaining Useful Life Prediction of Aircraft Engine," IEEE T. Cybernetics. 2021.

[11] R. Ren, T. Hung, and K. C. Tan, "A generic deep-learning-based approach for automated surface inspection," IEEE T. Cybernetics, vol. 48, no. 3, pp. 929-940. 2017.

[12] H. Shang, C. Sun, J. Liu, X. Chen, and R. Yan, "Deep learning-based borescope image processing for aero-engine blade in-situ damage detection," Aerosp. Sci. Technol., vol. 123, p. 107473. 2022.

[13] J. Aust, S. Shankland, D. Pons, R. Mukundan, and A. Mitrovic, "Automated defect detection and decision-support in gas turbine blade inspection," Aerospace, vol. 8, no. 2, p. 30. 2021.

[14] G. Yan-Ying, L. Zhi-Gang, and G. Qing-Ji, "Based on Weighted Morphology Aero-engine Bore Scope Cracks Image Segmentation," Proceedings of the 2019 11th International Conference on Machine Learning and Computing, 2019, pp. 451-455.

[15] K. Holak and W. Obrocki, "Vision-based damage detection of aircraft engine's compressor blades," Diagnostyka, vol. 22. 2021.

[16] F. Zou, "Review of aero-engine defect detection technology," 2020 IEEE 4th Information Technology, Networking, Electronic and Automation Control Conference (ITNEC), IEEE, 2020, pp. 1524-1527.

[17] W. Li et al., "A perspective survey on deep transfer learning for fault diagnosis in industrial scenarios: Theories, applications and challenges," Mech. Syst. Signal Pr., vol. 167, p. 108487. 2022.

[18] O. Fink, Q. Wang, M. Svensen, P. Dersin, W. Lee, and M. Ducoffe, "Potential, challenges and future directions for deep learning in prognostics and health management applications," Eng. Appl. Artif. Intel., vol. 92, p. 103678. 2020.

[19] C. Y. Wong, P. Seshadri, and G. T. Parks, "Automatic borescope damage assessments for gas turbine blades via deep learning," AIAA Scitech 2021 Forum, 2021, p. 1488.

[20] I. Ren, F. Zahiri, G. Sutton, T. Kurfess, and C. Saldana, "A Deep Ensemble Classifier for Surface Defect Detection in Aircraft Visual Inspection," Smart and Sustainable Manufacturing Systems, vol. 4, no. 1. 2020.

[21] F. Kähler, O. Schmedemann, and T. Schüppstuhl, "Anomaly detection for industrial surface inspection: application in maintenance of aircraft components," Procedia CIRP, vol. 107, pp. 246-251. 2022.

[22] M. Svensen, D. Hardwick, and H. Powrie, "Deep neural networks analysis of borescope images," Proceedings of the European Conference of the PHM Society, Utrecht, The Netherlands, 2018, pp. 3-6.

[23] X. Bian, S. N. Lim, and N. Zhou, "Multiscale fully convolutional network with application to industrial inspection," 2016 IEEE winter conference on applications of computer vision (WACV), IEEE, 2016, pp. 1-8.

[24] Z. Qu, C. Cao, L. Liu, and D. Zhou, "A deeply supervised convolutional neural network for pavement crack detection with multiscale feature fusion," IEEE T. Neur. Net. Lear. 2021.

[25] B. Hu et al., "A lightweight spatial and temporal multi-feature fusion network for defect detection," IEEE T. Image Process., vol. 30, pp. 472-486. 2020.

[26] Q. Zou, Z. Zhang, Q. Li, X. Qi, Q. Wang, and S. Wang, "Deepcrack: Learning hierarchical convolutional features for crack detection," IEEE T. Image Process., vol. 28, no. 3, pp. 1498-1512. 2018.

[27] J. Chen, Z. Liu, H. Wang, A. Núñez, and Z. Han, "Automatic defect detection of fasteners on the catenary support device using deep convolutional neural network," IEEE T. Instrum. Meas., vol. 67, no. 2, pp. 257-269. 2017.

[28] X. Ren, W. Lin, X. Yang, X. Yu, and H. Gao, "Data Augmentation in Defect Detection of Sanitary Ceramics in Small and Non-iid Datasets," IEEE T. Neur. Net. Lear. 2022.

[29] J. Liu, C. Wang, H. Su, B. Du, and D. Tao, "Multistage GAN for fabric defect detection," IEEE T. Image Process., vol. 29, pp. 3388-3400. 2019.

[30] X. Dong, C. J. Taylor, and T. F. Cootes, "Defect detection and classification by training a generic convolutional neural network encoder," IEEE T. Signal Proces., vol. 68, pp. 6055-6069. 2020.

[31] S. Ji, S. Pan, E. Cambria, P. Marttinen, and S. Y. Philip, "A survey on knowledge graphs: Representation, acquisition, and applications," IEEE T. Neur. Net. Lear., vol. 33, no. 2, pp. 494-514. 2021.

[32] J. Sun, W. Zheng, Q. Zhang, and Z. Xu, "Graph neural network encoding for community detection in attribute networks," IEEE T. Cybernetics. 2021.

[33] F. M. Bianchi, D. Grattarola, L. Livi, and C. Alippi, "Hierarchical representation learning in graph neural networks with node decimation pooling," IEEE T. Neur. Net. Lear. 2020.

[34] Z. Chen, J. Xu, T. Peng, and C. Yang, "Graph convolutional network-based method for fault diagnosis using a hybrid of

measurement and prior knowledge," IEEE T. Cybernetics. 2021.

[35] X. Zhao, J. Yao, W. Deng, P. Ding, J. Zhuang, and Z. Liu, "Multi-Scale Deep Graph Convolutional Networks for Intelligent Fault Diagnosis of Rotor-Bearing System Under Fluctuating Working Conditions," IEEE T. Ind. Inform. 2022.

[36] Z. Chen et al., "Graph neural network-based fault diagnosis: a review," arXiv preprint arXiv:2111.08185. 2021.

[37] A. Luo, X. Li, F. Yang, Z. Jiao, H. Cheng, and S. Lyu, "Cascade graph neural networks for RGB-D salient object detection," European Conference on Computer Vision, Springer, 2020, pp. 346-364.

[38] S. Chen, Z. Li, and Z. Tang, "Relation r-cnn: A graph based relation-aware network for object detection," IEEE Signal Proc. Let., vol. 27, pp. 1680-1684. 2020.

[39] S. Ren, K. He, R. Girshick, and J. Sun, "Faster r-cnn: Towards real-time object detection with region proposal networks," Advances in neural information processing systems, vol. 28. 2015.

[40] X. Tao, D. Zhang, Z. Wang, X. Liu, H. Zhang, and D. Xu, "Detection of power line insulator defects using aerial images analyzed with convolutional neural networks," IEEE Transactions on Systems, Man, and Cybernetics: Systems, vol. 50, no. 4, pp. 1486-1498. 2018.

[41] B. Su, H. Chen, P. Chen, G. Bian, K. Liu, and W. Liu, "Deep learning-based solar-cell manufacturing defect detection with complementary attention network," IEEE T. Ind. Inform., vol. 17, no. 6, pp. 4084-4095. 2020.

[42] K. Shih, C. Chiu, J. Lin, and Y. Bu, "Real-time object detection with reduced region proposal network via multi-feature concatenation," IEEE T. Neur. Net. Lear., vol. 31, no. 6, pp. 2164-2173. 2019.

[43] C. Song, W. Xu, G. Han, P. Zeng, Z. Wang, and S. Yu, "A cloud edge collaborative intelligence method of insulator string defect detection for power IIoT," IEEE Internet of Things Journal, vol. 8, no. 9, pp. 7510-7520. 2020.

[44] Z. Zeng, B. Liu, J. Fu, and H. Chao, "Reference-based defect detection network," IEEE T. Image Process., vol. 30, pp. 6637-6647. 2021.

[45] H. Zhang, M. Lin, G. Yang, and L. Zhang, "Escnet: An end-to-end superpixel-enhanced change detection network for very-high-resolution remote sensing images," IEEE T. Neur. Net. Lear. 2021.

[46] N. Lv, C. Chen, T. Qiu, and A. K. Sangaiah, "Deep learning and superpixel feature extraction based on contractive autoencoder for change detection in SAR images," IEEE T. Ind. Inform., vol. 14, no. 12, pp. 5530-5538. 2018.

[47] D. Tao, J. Cheng, M. Song, and X. Lin, "Manifold ranking-based matrix factorization for saliency detection," IEEE T. Neur. Net. Lear., vol. 27, no. 6, pp. 1122-1134. 2015.

[48] X. Zhou et al., "A surface defect detection framework for glass bottle bottom using visual attention model and wavelet transform," IEEE T. Ind. Inform., vol. 16, no. 4, pp. 2189-2201. 2019.

[49] K. Han, Y. Wang, J. Guo, Y. Tang, and E. Wu, "Vision GNN: An Image is Worth Graph of Nodes," arXiv preprint arXiv:2206.00272. 2022.

[50] G Li, M Muller, A Thabet, and Bernard Ghanem, "Deepgcns: Can gcns go as deep as cnns?" Proceedings of the IEEE/CVF International Conference on Computer Vision, 2021, pp. 9267-9276.

[51] B. Knyazev, G. W. Taylor, and M. Amer, "Understanding attention and generalization in graph neural networks," Advances in neural information processing systems, vol. 32. 2019.

[52] D. Wang, H. Xiao, and D. Wu, "Application of unsupervised adversarial learning in radiographic testing of aeroengine turbine blades," NDT E Int., vol. 134, p. 102766. 2023.

[53] D. Wang, H. Xiao, and S. Huang, "Automatic Defect Recognition and Localization for Aeroengine Turbine Blades Based on Deep Learning," Aerospace, vol. 10, no. 2, p. 178. 2023.

[54] B. E. Jaeger, S. Schmid, C. U. Grosse, A. Gögelein, and F. Elischberger, "Infrared Thermal Imaging-Based Turbine Blade Crack Classification Using Deep Learning," J. Nondestruct. Eval., vol. 41, no. 4, p. 74. 2022.

[55] Y. Kim and J. Lee, "Videoscope-based inspection of turbofan engine blades using convolutional neural networks and image processing," Structural Health Monitoring, vol. 18, no. 5-6, pp. 2020-2039. 2019.

[56] Z. Shen, X. Wan, F. Ye, X. Guan, and S. Liu, "Deep learning based framework for automatic damage detection in aircraft engine borescope inspection," 2019 International Conference on Computing, Networking and Communications (ICNC), IEEE, 2019, pp. 1005-1010.

[57] X. Li, W. Wang, L. Sun, B. Hu, L. Zhu, and J. Zhang, "Deep learning-based defects detection of certain aero-engine blades and vanes with DDSC-YOLOv5s," Sci. Rep., vol. 12, no. 1, p. 13067. 2022.

[58] L. Chen, L. Zou, C. Fan, and Y. Liu, "Feature weighting network for aircraft engine defect detection," International Journal of Wavelets, Multiresolution and Information Processing, vol. 18, no. 03, p. 2050012. 2020.

[59] C. Y. Wong, P. Seshadri, and G. T. Parks, "Automatic borescope damage assessments for gas turbine blades via deep learning," AIAA Scitech 2021 Forum, 2021, p. 1488.

[60] S. Liu, L. Qi, H. Qin, J. Shi, and J. Jia, "Path aggregation network for instance segmentation," Proceedings of the IEEE conference on computer vision and pattern recognition, 2018, pp. 8759-8768.

[61] J. Pang, K. Chen, J. Shi, H. Feng, W. Ouyang, and D. Lin, "Libra r-cnn: Towards balanced learning for object detection," Proceedings of the IEEE/CVF conference on computer vision and pattern recognition, 2019, pp. 821-830.

[62] P. Sun et al., "Sparse r-cnn: End-to-end object detection with learnable proposals," Proceedings of the IEEE/CVF conference on computer vision and pattern recognition, 2021, pp. 14454-14463.

[63] Z. Liu et al., "Swin transformer: Hierarchical vision transformer using shifted windows," Proceedings of the IEEE/CVF International Conference on Computer Vision, 2021, pp. 10012-10022.